\documentclass[10pt,twocolumn,letterpaper]{article}

\usepackage[pagenumbers]{cvpr} 
\usepackage{multirow}

\usepackage[dvipsnames]{xcolor}

\usepackage{epsfig}
\usepackage{graphicx}
\usepackage{amsmath}
\usepackage{amssymb}
\usepackage{multirow}
\usepackage{graphics}
\usepackage{threeparttable}
\usepackage{color}
\usepackage[normalem]{ulem}
\usepackage{float}
\usepackage{amsfonts}
\usepackage{bm}
\usepackage{bbm}
\usepackage{enumitem}
\usepackage{algorithmic,algorithm}
\usepackage{natbib}
\usepackage{subcaption}
\usepackage{algorithm}
\usepackage{array}
\usepackage{colortbl}
\usepackage{pifont}
\usepackage{booktabs}
\usepackage{mathtools}
\usepackage{siunitx}
\usepackage{caption} 
\usepackage{xcolor}
\usepackage{subcaption} 
\usepackage[nopar]{lipsum}
\usepackage{listings}
\usepackage[export]{adjustbox}
\usepackage{makecell}

\usepackage{mathtools}
\usepackage{siunitx}
\usepackage[nopar]{lipsum}
\usepackage{listings}

\usepackage{xargs}                      
\usepackage{xcolor}
\usepackage{wrapfig} 
\definecolor{myy}{RGB}{126,95,0}
\definecolor{mygray}{gray}{.9}
\definecolor{Gray}{gray}{0.9}
\definecolor{bblue}{RGB}{30,80,120}
\definecolor{mygray1}{gray}{.7}
\definecolor{ggray}{RGB}{127,127,127}
\definecolor{defaultcolor}{gray}{.9}
\definecolor{dark-gray}{gray}{0.20}
\newcommand{\reshl}[2]{
	\textbf{#1} \fontsize{7.5pt}{1em}\selectfont\color{mygreen}{$\uparrow$ \textbf{#2}}
}
\newcommand{\deshl}[2]{
	{#1} \fontsize{7.5pt}{1em}\selectfont\color{mygreen}{$\uparrow$ \textbf{#2}}
}
\definecolor{mygreen}{HTML}{39b54a}

\newcolumntype{x}[1]{>{\centering\arraybackslash}p{#1pt}}
\newcolumntype{y}[1]{>{\raggedright\arraybackslash}p{#1pt}}
\newcolumntype{z}[1]{>{\raggedleft\arraybackslash}p{#1pt}}

\newlength\savewidth

\definecolor{cvprblue}{rgb}{0.21,0.49,0.74}
\usepackage[pagebackref,breaklinks,colorlinks,citecolor=cvprblue]{hyperref}

\def\paperID{16725} 
\def\confName{CVPR}
\def\confYear{2024}

\title{Multimodal Pathway: Improve Transformers with Irrelevant Data from Other Modalities}

\author{
	{Yiyuan Zhang}$^{1}$
	\quad {Xiaohan Ding}$^{2}$
	\quad  {Kaixiong Gong}$^{1}$ 
	 \quad {Yixiao Ge}$^{2}$
        \quad {Ying Shan}$^{2}$
	\quad {Xiangyu Yue}$^{1}$\thanks{Corresponding Author} \\
	\textsuperscript{1}MMLab, The Chinese University of Hong Kong
        \quad
	\textsuperscript{2}Tencent AI Lab~~~ \\
	{\tt\small yiyuanzhang.ai@gmail.com, xiaohding@gmail.com, xyyue@ie.cuhk.edu.hk} \\
 \url{https://ailab-cvc.github.io/M2PT/}
}

\begin{document}
\maketitle
\begin{abstract}
We propose to improve transformers of a specific modality with irrelevant data from other modalities, \eg, improve an ImageNet model with audio or point cloud datasets. We would like to highlight that the data samples of the target modality are irrelevant to the other modalities, which distinguishes our method from other works utilizing paired (\eg, CLIP) or interleaved data of different modalities. We propose a methodology named Multimodal Pathway - given a target modality and a transformer designed for it, we use an auxiliary transformer trained with data of another modality and construct pathways to connect components of the two models so that data of the target modality can be processed by both models. In this way, we utilize the universal sequence-to-sequence modeling abilities of transformers obtained from two modalities. As a concrete implementation, we use a modality-specific tokenizer and task-specific head as usual but utilize the transformer blocks of the auxiliary model via a proposed method named Cross-Modal Re-parameterization, which exploits the auxiliary weights without any inference costs. On the image, point cloud, video, and audio recognition tasks, we observe significant and consistent performance improvements with irrelevant data from other modalities. The code and models are available at \url{https://github.com/AILab-CVC/M2PT}.
\end{abstract}

\section{Introduction} ~\label{sec:intro}

Transformers~\cite{vaswani2017attention,dosovitskiy2020image,touvron2021training,ge2023advancing} are widely adopted in various tasks across modalities, such as text classification~\citep{devlin2018bert}, object detection~\citep{carion2020end}, point cloud analysis~\citep{zhao2021pointtransformer}, and audio spectrogram recognition~\citep{gong2021ast}.
Apart from numerous unimodal tasks, transformers are also effective on multimodal data, \eg, CLIP~\citep{radford2021learning} uses image-text pairs to achieve superior performance in image recognition. Transformers' success in multiple modalities demonstrates their abilities to universally establish sequence-to-sequence modeling, given the input sequences (\textit{i.e.}, tokens) which can be seen as the universal embeddings of data of multiple modalities~\citep{dosovitskiy2020image,carion2020end,zhao2021pointtransformer,gong2021ast,zhang2023meta}. For brevity, we refer to such ability as the \emph{universal modeling ability}.

    \begin{figure*}[t]
        \begin{center}
        \includegraphics[width=0.88\linewidth]{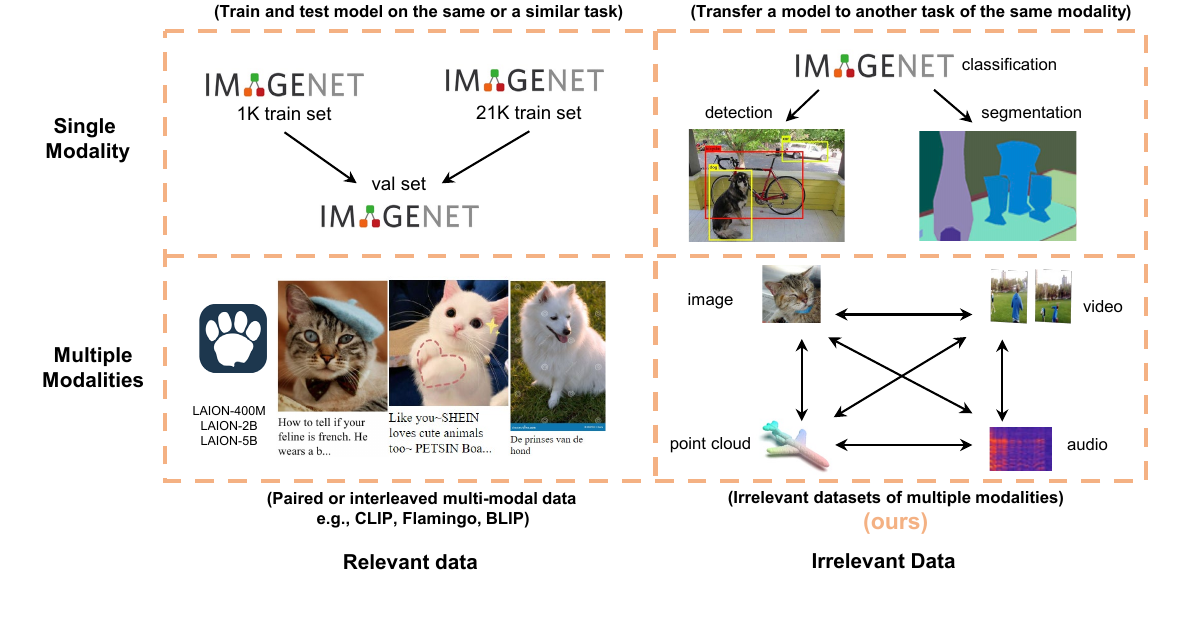}
        \vspace{-3mm}
        \caption{Compared to the known paradigms which use well-aligned multimodal data, we focus on scenarios where the data samples are from multiple modalities but irrelevant, which is an open problem in the literature.}
        \label{fig:motivation}
        \end{center}
        \vskip -0.1in
    \end{figure*}

We would like to note that CLIP~\citep{radford2021learning} represents the significant success of a methodology that improves a model's performance on a certain modality (\textit{i.e.}, image) with the help of data from another modality (\textit{i.e.}, text), but the limitation is also apparent - \textbf{the data samples from the two modalities must be relevant} (\eg, paired, in this case). This limitation seems so inevitable that it hardly attracts research interest from the literature. Taking another two modalities, image and audio, as an example, we may expect that training with image-audio pairs may help the model recognize images (if we build a dataset with enough image-audio pairs and re-design the model to use the audio labels as the supervision, just like CLIP does with image-text pairs), but \textbf{it seems hard to believe that a pure audio dataset would improve a model's performance on ImageNet classification without any relevance between the audio and image samples}.

In this paper, we propose to improve a transformer's performance on a certain modality even with irrelevant data from another modality, as shown in Figure~\ref{fig:motivation}. The motivation is that we can see a training process on a certain modality as converting the data of the modality to sequences (\textit{i.e.}, tokens) and establishing sequence-to-sequence mappings with the transformer blocks. For a specific modality, we reckon that the trained model has knowledge encoded in the sequence-to-sequence modeling that can facilitate another modeling process whose input sequences are obtained from another modality. In other words, apart from the obvious modality-specific knowledge acquired through training on a specific modality, we seek the \textbf{modality-complementary knowledge of sequence-to-sequence modeling in transformers} and will show that \textbf{it does exist}.

However, given a target modality, it seems difficult to design the model to utilize some irrelevant data of another modality because the data samples of different modalities (\eg, image and audio) may vary significantly in the semantics, data format, preprocessing, and it seems hardly possible to design a reasonable objective function since there is no relevance between any two samples. In this paper, we solve this problem by not directly mixing training data of two modalities but \emph{seeing a model trained on a specific unimodal dataset as a proxy of the corresponding modality and using the model instead}. Specifically, given a target modality and an auxiliary modality, we propose a framework named \emph{Multimodal Pathway} to improve the performance on the target modality by \emph{using two transformers respectively trained with the unimodal data of the two modalities}. We construct \emph{pathways} across the components of the target and auxiliary models to exploit the modality-complementary knowledge encoded in the latter to help the former. Note pathway is an abstract concept that may refer to any connection between the two models. We name the model as \textbf{M}ulti\textbf{m}odal \textbf{P}athway \textbf{T}ransformer (\textbf{M2PT}) for brevity.

\begin{figure*}[t]
		\begin{center}
			\includegraphics[width=0.98\linewidth]{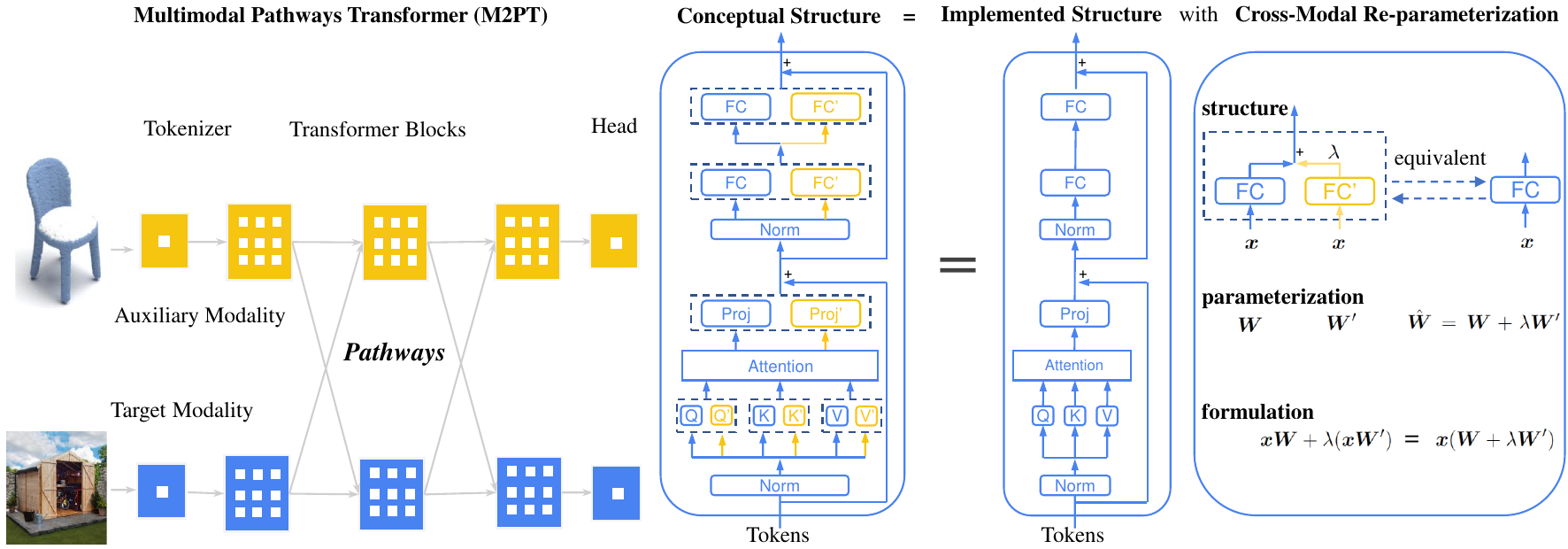}
			\caption{(\textbf{Left}) Framework of Multimodal Pathway Transformer (M2PT). We use point cloud and image modalities as an example. Common practices with transformers follow the same pipeline: using 1) tokenizers to convert the input data to sequences, 2) transformer blocks to process the sequences, and 3) heads to decode the sequences. We upgrade the sequence-to-sequence modeling by establishing \emph{pathways} between the components of different modalities so processing the tokens of a specific modality can utilize the transformer blocks trained with another modality. (\textbf{Middle}) Conceptual design of M2PT, where the pathways are implemented by letting a linear layer (including the Query/Key/Value/projection layers in the attention block and those in the FFN block) in the target model cooperate with its counterpart in the auxiliary model. (\textbf{Right}) Cross-Modal Re-parameterization efficiently realizes M2PT by re-parameterizing the weights of the target model with those of the auxiliary model, introducing marginal training costs and completely no inference costs.}
			\label{fig:framework}
			\vspace{-0.2in}
		\end{center}
	\end{figure*}
 
    \begin{figure}[ht]
        \centering
        \includegraphics[width=0.80\linewidth]{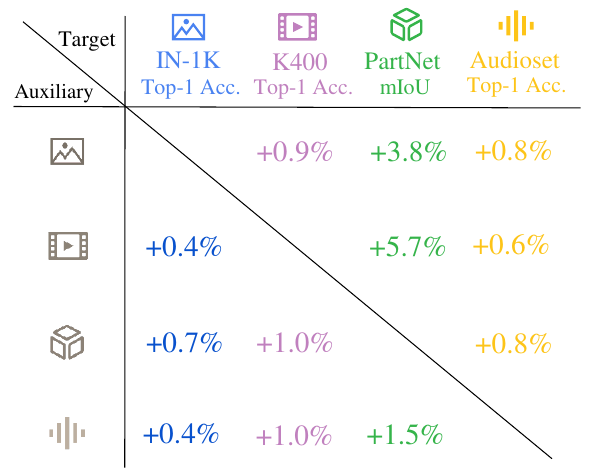}
    \vspace{-0.1in}
    \caption{Consistent improvements brought by M2PT across each pair of four modalities - image, video, point cloud, and audio. The metrics are ImageNet-1K accuracy, Kinetics-400 accuracy, PartNet mIoU, and AudioSet accuracy, respectively. The numbers represent the percentage of improvement of M2PT models relative to the performance of baseline models that are pretrained with MAE-style methods~\citep{he2022masked,pang2022masked,huang2022masked,zhou2022audio} on the four modalities, respectively.}
    \label{fig:summary}
    \vspace{-0.1in}
    \end{figure}

This paper proposes a simple yet effective implementation of M2PT, where the key is the concrete implementation of pathways that connect the two models. As discussed above, thanks to the universal modeling ability, transformers on different modalities may have different tokenizers, but their main bodies (\ie, transformer blocks) may have the same structure.~\footnote{Except for transformers, a recent work, UniRepLKNet~\cite{ding2023unireplknet}, reveals ConvNets also effectively handle embeddings extracted from different modalities with the same architecture (akin to transformers universally tokenizing and processing data of multiple modalities), achieving state-of-the-art performances in tasks including global weather forecasting and audio recognition.} For a target model and an auxiliary model with the same structure as the main bodies, a layer in the main body of the former should have a counterpart in the latter. For example, the counterpart of the Query layer in the 9th block of the target model, which is the 9th Query layer in the auxiliary model, should exist, and they play a similar role in the two models. Considering this, we build the connections between the two models by augmenting every linear layer in the transformer blocks of the target model with its counterpart in the auxiliary model. In such a conceptual design, we let the two layers take the same inputs and add up their outputs, as shown in Figure~\ref{fig:framework} (middle).

However, considering the budget on compute and latency, we desire an implementation of the Multimodal Pathway that realizes the pathways and makes good use of the auxiliary model but \emph{brings only marginal training cost and completely no inference cost}, compared to a regular model trained on the target modality. We note that the conceptual structure described above can be equivalently implemented by a re-parameterization method, which equivalently converts the connections between model structures (\textit{i.e.,} linear layers) into connections between the two models' weights. Specifically, we construct a pathway for each target linear layer by adding the corresponding weights of its counterpart in the trained auxiliary model scaled by a learnable multiplier that indicates the strength of the pathway, so that the method is named \emph{Cross-Modal Re-parameterization}. A significant strength of re-parameterization is that the extra training costs are marginal (\ie, the re-parameterized model will have the same number of linear layers as the original model, and each linear layer merely needs to compute the sum of two weight matrices before projecting the inputs) and we can merge the weights after training so that the structure and number of parameters of the resultant model will be identical to a regular model.

We experimented with the image, video, point cloud, and audio modalities. Figure~\ref{fig:summary} shows the relative improvements M2PT consistently brings among four modalities. Such results reveal that the modality-complementary knowledge of sequence-to-sequence modeling in transformers does exist. As an early exploration, our empirical studies confirm that such improvements are not solely due to the more parameters, and suggest that such modality-complementary knowledge may be related to the ability to generally process hierarchical representations. Abstraction hierarchy exists in multiple modalities with concepts ranging from low-level to high-level, which may explain the universality of the learned knowledge. In other words, as a transformer is being trained with images, it learns both (ability A) how to understand images and (ability B) how to generally transform the tokens from the lower-level patterns to a higher level without assuming they originally come from images. Meanwhile, as another transformer is being pretrained with audio data, it learns both a different ``ability A'' for audio and a similar ``ability B'', so that it can help the aforementioned transformer in image recognition.

In summary, our contributions are as follows:
\begin{itemize}[itemsep=0pt] 
    \item We propose Multimodal Pathway, which is a framework to improve transformers via exploiting models trained on other modalities.
    \item We propose an inference-cost-free implementation of Multimodal Pathway, which is named Cross-Modal Re-parameterization.
    \item Multimodal Pathway represents an early exploration in this direction, which offers a novel perspective. We realize significant and consistent improvements in four representative modalities, which demonstrates the potential of our method as a promising approach.
\end{itemize}

\section{Related Work} ~\label{sec:related}
\noindent\textbf{Unimodal pretraining}. The evolution of unimodal pretraining paradigms has transitioned from supervised to self-supervised paradigms. For instance, \citet{devlin2018bert} introduced the mask-reconstruction paradigm and achieved remarkable outcomes. At that time, visual pretraining largely emphasized contrastive learning~\citep{chen2020simple,he2020momentum,caron2021emerging}. Subsequently, leveraging the vast amounts of unlabeled data, the BERT paradigm gained traction and pioneers like MAE~\citep{he2022masked} successfully applied it to visual pretraining, while others \citep{pang2022masked,gong2021ast,tong2022videomae,zhang2023meta} extended this paradigm to areas like point cloud, audio, and video perception. 

We use MAE-style unimodal pretraining methods to obtain the weights on each modality for simplicity. We do not use supervised pretraining because we would like to ensure that two unimodal datasets are completely irrelevant by avoiding using labels, considering that the labels of two datasets may somehow overlap.

\noindent\textbf{Multimodal pretraining}. Existing multimodal learning methods require paired~\cite{wang2021vlmo,wang2021simvlm,zhu2023vl,han2023onellm} or interleaved data~\cite{alayrac2022flamingo}. In either case, the data samples of different modalities are well-aligned (\ie, strongly related). A recent study highlighted a main trend in the literature - \emph{existing multimodal pretraining methods are overly dependent on the well-aligned multimodal sample pairs/tuples}~\cite{xu2023multimodal}. For instance, VideoBERT~\cite{sun2019videobert} and CBT~\cite{sun2019learning} utilize well-aligned video and speech data;  

Nowadays, using the weakly-aligned or unpaired/unaligned multimodal data as the pretraining corpora remains understudied~\cite{xu2023multimodal}. This work represents an early exploration in this direction, which serves to fill this gap in the field and contributes to multimodal  calibration~\cite{wang2020makes}.

\noindent\textbf{Structural Re-parameterization} is a methodology that constructs extra structures (\eg, convolutional layers) during training and converts the trained structures via transforming the parameters~\cite{ding2021repvgg,ding2022scaling,ding2023unireplknet}. A primary drawback of Structural Re-parameterization is that the constructed layers must participate in the computations with the inputs, resulting in significant extra training costs.

In contrast, Cross-Modal Re-parameterization is a simple re-parameterization method that is more efficient than Structural Re-parameterization. Specifically, the extra computation of each re-parameterized layer in the forward computation adds up two weight matrices,

\section{Method} ~\label{sec:method}
\subsection{Architectural Design}

We design a transformer for a specific modality as three modules - the modality-specific tokenizer, the modality-agnostic transformer blocks, and the modality-specific head. We assume the dimension of tokens is $D$, which is a pre-defined architectural hyper-parameter, and describe how to tokenize the input data of multiple modalities into $D$-dimensional tokens.

\noindent\textbf{Image tokenizer}. We represent an image by \( \boldsymbol{x}_I \in \mathbb{R}^{H \times W \times C} \), where \( (H, W) \) specifies the image's resolution, and \( C \) is the number of channels. With an image patch of $(S,S)$, we obtain:
    \begin{equation}
        \boldsymbol{x}_I \in \mathbb{R}^{ H \times W \times C } \rightarrow \boldsymbol{x}_I^{\prime} \in \mathbb{R}^{\frac{HW}{S^2} \times D }\,.
        \label{eq:image:token}
    \end{equation}
    \textbf{Video tokenizer}. Analogous to 2D images, we use video patches as the basic units for learning video representations. Given an $N$-frame video \(\bm{x} \in \mathbb{R}^{N \times H \times W \times C }\), similar to images, we use an $S$$\times$$S$ embedding layer so that
    \begin{equation}
        \label{eq:video:token}
        \boldsymbol{x}_V\in \mathbb{R}^{ N\times H \times W \times C } \rightarrow \boldsymbol{x}_V^{\prime} \in \mathbb{R}^{\frac{NHW}{S^2} \times D }\,.
    \end{equation}
    Following ViT~\cite{dosovitskiy2020image}, we use $S=16$ by default.
    
	\noindent\textbf{Point cloud tokenizer}. Given a point cloud \(\mathcal{X} = \{\boldsymbol{x}_i\}_{i=1}^P\) comprising \(P\) points, each point \(\boldsymbol{x}_i\) is defined as \(\boldsymbol{x}_i = (\boldsymbol{p}_i, \boldsymbol{f}_i)\), where \(\boldsymbol{p}_i \in \mathbb{R}^3\) denotes the 3D coordinates and \(\boldsymbol{f}_i \in \mathbb{R}^c\) encodes the attributes, \eg, color, viewpoint, normal, \etc. We use the Farthest Point Sampling to sample a representative skeleton from the original points at a fixed sampling ratio of 1/4, then \(K\)-Nearest Neighbor method to group proximate points. Then we model the geometric relevance by constructing an adjacency matrix $\mathbb{R}^{\frac{P}{4} \times \frac{P}{4}}$ between each pair of groups, which is then projected into $D$-dimensional tokens. That is
	\begin{equation}
		\boldsymbol{x}_P \in \mathbb{R}^{ P \times (3+c)} \rightarrow \boldsymbol{x}_P^\prime \in \mathbb{R}^{\frac{P}{4} \times \frac{P}{4} } \rightarrow \boldsymbol{x}_P^{\prime\prime} \in \mathbb{R}^{ \frac{P}{4} \times D }\,.
		\label{eq:pcd:token}
	\end{equation}
     \textbf{Audio spectrogram tokenizer}. Let $T$ and $F$ be the numbers of time frames and frequency bins, we use $\boldsymbol{x}_A \in \mathbb{R}^{T\times F}$ to represent a sample. Analogous to 2D images, we see an audio sample as a single-channel image and use a similar embedding layer so that
	\begin{equation}
		\boldsymbol{x}_A \in \mathbb{R}^{T \times F} \rightarrow \boldsymbol{x}_A^\prime \in \mathbb{R}^{\frac{TF}{S^2} \times D}.
		\label{eq:audio:token}
	\end{equation}
    In our AudioSet experiments, we have $T$=$F$=128, $S$=16.

\noindent\textbf{Transformer blocks}. We adopt the structural design of the transformer blocks in Vision Transformer (ViT)~\citep{dosovitskiy2020image}, where each transformer block comprises a self-attention block and a Feed-Forward Network (FFN) block. The linear layers include the Query/Key/Value/projection layers in the attention block and two layers in the FFN block. For fairness and reproducibility, we use the same architectural hyper-parameters (\eg, dimension of tokens, number of blocks, and number of heads) as ViT-Base for every M2PT model on every modality.

\subsection{Cross-Modal Re-parameterization}

For an M2PT model on a specific modality, we use Cross-Modal Re-parameterization in the transformer blocks to utilize another model's weights trained on another modality. Specifically, let $\theta$ be an arbitrary trainable parameter of a layer in the transformer, $x$ be the input, and $y$ be the output, we use $f$ to denote the operation so that $y=f(x;\theta)$. With Cross-Modal Re-parameterization, we simply re-parameterize the layer with parameters of its counterpart in another modal that is trained on another modality. Let $\theta^\prime$ be the parameter of the counterpart, the operation becomes
\begin{equation}
    y=f(x;\theta + \lambda\theta^\prime) \,.
    \label{eq:rep}
\end{equation}
We refer to $\lambda$ as the \emph{Cross-Modal Scale} and $\theta^\prime$ as the \emph{Cross-Modal Parameter}. After training, we merge the model by computing and saving $\hat{\theta} = \theta + \lambda\theta^\prime$ so that the model will no longer have extra parameters and the inference costs and model size will be identical to a regular model.

With Cross-Modal Re-parameterization, we equivalently realize the proposed M2PT transformer block with marginal training costs and completely no inference costs. For a linear layer whose parameters form a matrix $\boldsymbol{W}\in\mathbb{R}^{D_{in}\times D_{out}}$ and the inputs and outputs are matrices $\boldsymbol{x}\in\mathbb{R}^{B\times D_{in}}$ and $\boldsymbol{y}\in\mathbb{R}^{B\times D_{out}}$. We omit the bias term for brevity and the original operation is formulated by
\begin{equation}
    \boldsymbol{y} = \boldsymbol{x}\boldsymbol{W} \,.
\end{equation}
As described in the conceptual structure depicted in Figure~\ref{fig:framework}, the linear layer and its counterpart take the same input. The output will be 
\begin{equation}
    \boldsymbol{y} = \boldsymbol{x}\boldsymbol{W} + \lambda(\boldsymbol{x}\boldsymbol{W}^\prime) \,.
\end{equation}
Note
\begin{equation}
    \boldsymbol{x}\boldsymbol{W} + \lambda(\boldsymbol{x}\boldsymbol{W}^\prime) = \boldsymbol{x}(\boldsymbol{W} + \lambda\boldsymbol{W}^\prime)\,,
\end{equation}
so that the two layers can be equivalently implemented by a single layer that has a trainable scalar $\lambda$ and an additional trainable matrix which is initialized with the counterpart in the auxiliary model. Both the original weight matrix and the additional one are trainable. At each forward computation, the layer computes the equivalent weight matrix and then uses it to project the input, which is
\begin{equation}\label{eq-reparam}
    \boldsymbol{y} = \boldsymbol{x}(\boldsymbol{W} + \lambda\boldsymbol{W}^\prime) \,.
\end{equation}
After training, we merge the parameters by computing $\hat{\boldsymbol{W}}=\boldsymbol{W} + \lambda\boldsymbol{W}^\prime$ and save it only. For inference, we simply construct a regular linear layer and load $\hat{\boldsymbol{W}}$.

In summary, to construct and use an M2PT with Cross-Modal Re-parameterization, we
\begin{itemize}[itemsep=0pt] 
    \item Construct the tokenizer and head according to the target modality.
    \item Construct the transformer blocks with Cross-Modal Re-parameterization. For each linear layer, except for the original weight matrix, we add an extra trainable weight matrix and initialize it with the corresponding one from a transformer trained on the auxiliary modality, and add a trainable scalar parameter initialized with 0.
    \item Train the re-parameterized cross-modal model just like we train a regular model.
    \item After training, convert the trained model and save the converted one for inference.
\end{itemize}

\section{Experiments} ~\label{sec:exp}
\subsection{Setup}\label{sec:exp:setup}
\noindent\textbf{Datasets}. For image recognition, we evaluate the models' performance on three representative image datasets. 1) ImageNet-1K~\citep{deng2009imagenet} contains nearly 1.3 million images of 1000 categories. 2) MSCOCO 2017~\citep{lin2014microsoft} is a common benchmark for object detection. M2PT is trained on the \texttt{train} set and evaluated on the \texttt{val} set with Mask RCNN~\citep{he2017mask}. 3) ADE-20K~\citep{zhou2017scene} is used for semantic segmentation experiments with UperNet~\citep{xiao2018unified} and we adopt the single-scale evaluation setting. For point cloud, we evaluate the performance of M2PT on ShapeNetPart~\citep{yi2016scalable}, which contains 16,880 models and 16 categories. For audio recognition, following AudioMAE~\citep{huang2022masked}, we utilize the AudioSet-2k~\citep{gemmeke2017audio} dataset. For video, we experiment on the action recognition dataset, Kinetics-400~\citep{kay2017kinetics}, which contains 240k training videos and 20k validation videos from 400 classes.

\begin{table*}[t]
\caption{\textbf{Experimental results on image recognition tasks.} On ImageNet, we report the results with the linear layers in transformer blocks finetuned (tune acc) or fixed (fix acc). The architecture of every model is ViT-B. The relative improvements over the baselines are shown in \color{mygreen}{green}.  * The standard error of M2PT on image recognition tasks is 0.04.}
\centering
\tabcolsep 7pt
\small
\begin{tabular}{llllll}
\toprule
 \multirow{2}{*}{\bfseries Method} & \multicolumn{2}{c}{ImageNet} & \multicolumn{2}{c}{MS COCO} & ADE20K \\
& tune acc(\%) & fix acc(\%) & $\text{AP}_{box}$(\%) & $\text{AP}_{mask}$(\%) & mIOU(\%) \\
\midrule
\multicolumn{6}{@{\;}l}{\bf Pretrained setting} \\
\quad SemMAE\citep{li2022semmae} &83.4&65.0&- & - & 46.3\\
\quad {MFF}~\citep{liu2023improving} & 83.6& 67.0 & 48.1 & 43.1   & 47.9 \\
\hline
\quad MAE$^{*}$\citep{he2022masked}  & 83.3 & 65.6 & 47.3 & 42.4 & 46.1 \\
\quad M2PT-Video~(Ours)  & \reshl{83.6}{0.4\%} & \reshl{67.1}{2.3\%} & - & - & - \\
\quad M2PT-Audio~(Ours)  & \reshl{83.7}{0.4\%} & \reshl{67.3}{2.6\%} & - & - & - \\
\quad M2PT-Point~(Ours)  & \reshl{83.9}{0.7\%} & \reshl{67.8}{3.4\%} & \reshl{50.0}{5.7\%} & \reshl{44.0}{3.8\%} & \deshl{47.9}{3.9\%} \\
\midrule
\multicolumn{6}{@{\;}l}{\bf From-scratch setting} \\
\quad ViT~\citep{dosovitskiy2020image} & 76.5 & 14.5 &46.2 & 40.5 & 39.7 \\
\quad M2PT-Point~(Ours)  & \reshl{81.9}{7.1\%} & \reshl{19.5}{34.5\%} & \reshl{48.9}{5.8\%} &  \reshl{42.2}{4.2\%} & \reshl{42.5}{7.1\%} \\
\bottomrule
\end{tabular}
\label{tab:image}
\end{table*}

\noindent\textbf{Experimental details}. For a pair of target modality and auxiliary modality, we obtain the auxiliary model by self-supervised training on a dataset of the auxiliary modality. Specifically, the auxiliary image model is pretrained with MAE~\citep{he2022masked} on ImageNet-1K~\citep{deng2009imagenet}, the auxiliary point cloud model is pretrained with Point-MAE~\citep{pang2022masked} on ShapeNet~\citep{chang2015shapenet}, the auxiliary audio model is pretrained with AudioMAE~\citep{huang2022masked} on AudioSet-2M~\citep{gemmeke2017audio}, the auxiliary video model is pretrained with VideoMAE~\citep{tong2022videomae} on Kinetics-700~\citep{kay2017kinetics}. We do not use supervised pretraining because we would like to eliminate the effects of labels in the pretraining datasets so that we can ensure the irrelevance of the data samples, considering that the labels of two datasets may somehow overlap. In terms of the initialization of the target model, we adopt two settings. 1) The target model (\textit{i.e.}, the parameters denoted by $\boldsymbol{W}$ in Eq.~\ref{eq-reparam}) is initialized with the aforementioned weights pretrained with the self-supervised methods on the target modality. We finetune the M2PT model with the default finetuning configurations described by the corresponding pretraining methods. The baseline model is also initialized with the pretrained weights and fine-tuned with identical configurations so that this setting is referred to as the \emph{pretrained setting} for brevity. 2) The target model is initialized as usual, and we use the widely adopted training configurations to train the M2PT model. The baseline model is trained from scratch with identical configurations for fair comparisons so that the setting is referred to as the \emph{from-scratch setting} for brevity.

\noindent\textbf{Metrics}. We report the performance of M2PT models on various datasets, including top-1 accuracy for ImageNet-1K, AudioSet, Kinetics-400, mIoU for ADE20K, ShapeNetPart and PartNet, and box/mask AP for MS COCO. To fairly assess the performance improvements over the baselines in multiple metrics, we also report the relative percentage of improvement in Table~\ref{tab:image},~\ref{tab:3d},~\ref{tab:audio}, and~\ref{tab:video}.

\subsection{Main Results} ~\label{sec:exp:main_results}

\noindent\textbf{Image recognition.} We first conduct a group of experiments under the pretrained setting, where the target weights are initialized with a ViT pretrained with MAE on ImageNet, and the auxiliary weights are from the models pretrained on video, audio, and point datasets, respectively. Such three models, which are labeled as M2PT-Video, M2PT-Audio, and M2PT-Point, respectively, and the baseline (the original MAE-pretrained ViT) are trained on ImageNet with the finetuning configurations originally adopted by MAE~\citep{he2022masked}, and the resultant accuracies are reported in the ``tune acc'' column in Table~\ref{tab:image}. Then we transfer the best-performing model, which is M2PT-Point, to COCO object detection and ADE20K semantic segmentation tasks. The improvements are significant: the ImageNet accuracy improves from 83.3 to 83.9, the COCO box AP improves from 47.3 to 50.0, and the ADE20K mIoU improves from 46.1 to 47.9, so the relative improvements are 0.7\%, 5.7\%, and 3.9\%, respectively. 

Apart from finetuning the target and auxiliary weights, we test another setting where the parameters of linear weights in transformer blocks are fixed, and only the Cross-Modal Scales together with the classifier are trainable. The accuracies are reported in the ``fix acc'' column. Naturally, under this setting, the baseline should be the MAE-pretrained ViT where only the classifier is trainable. Impressively, the relative improvement becomes more significant (65.6$\to$67.8 so that the relative improvement is 3.4\%), demonstrating that the weights obtained from the auxiliary modality work on another modality, even if the weights are fixed. We would like to note MAE is a powerful pretraining method, and it is challenging to gain further improvements on top of MAE. Some insightful recent methods~\cite{li2022semmae,liu2023improving} improved MAE but our results are more significant.

On the other hand, under the from-scratch setting, the baseline is a ViT trained from scratch, and the target weights of M2PT are also randomly initialized. The accuracy is drastically improved from 81.9 to 76.5 so the relative improvement is 7.1\%, suggesting the auxiliary weights significantly facilitate the training process. Intuitively, the Cross-Modal Scales are initialized with 0 but will soon become non-zero as the training proceeds so the model will be gradually influenced by the auxiliary weights and benefit from the modality-complementary knowledge. When we transfer such two models to COCO and ADE20K, we observe consistent improvements in the box AP and mIoU.

\noindent\textbf{3D point cloud understanding.} Table~\ref{tab:3d} presents the experimental results on ShapeNetPart and PartNet datasets, where we compare M2PT with existing point cloud pretraining methods such as Point-BERT~\citep{pang2022masked} and Point-MAE~\citep{yu2022pointbert}. M2PT consistently improves the class mIoU from 84.2 to 85.6 and instance mIoU from 86.1 to 87.5 on ShapeNetPart and raises the mIoU from 47.4 to 50.1 on PartNet. Under the from-scratch setting, we also observe consistent improvements.

\noindent\textbf{Audio recognition.} For the pretrained setting, the target weights are initialized with an AudioMAE-pretrained model. As shown in Table~\ref{tab:audio}, we compare M2PT with existing competitive methods including SSAST~\citep{gong2022ssast}, AST~\citep{gong2021ast}, and AudioMAE~\citep{huang2022masked}. M2PT improves the top-1 accuracy by 0.8\% relatively on the Audioset balanced split, demonstrating that M2PT is also effective in audio recognition. Under the from-scratch setting, M2PT brings out a relative improvement of 3.6\%. 

\begin{table}[t]
		\caption{\textbf{Experimental results on point cloud datasets}. We report the class mIoU ($\text{mIoU}_{C}$) and instance $\text{mIoU}_{I}$ on ShapeNetPart and mIoU on PartNet. The relative improvements over the baselines are shown in \color{mygreen}{green}.}
  \vspace{-0.1in}
  \label{tab:3d}
		\centering	
  \small
  \resizebox{0.95\linewidth}{!}{
        \begin{tabular}{lccc}
        \toprule
        \multirow{2}{*}{\bfseries Method} & \multicolumn{2}{c}{ShapeNetPart} & PartNet\\
        \cline{2-4} & $\text{mIoU}_{C}$ (\%) &  $\text{mIoU}_{I}$ (\%)
        & mIoU (\%) \\ \hline 
        \multicolumn{4}{@{\;}l}{\bf Pretrained setting} \\
        \quad PointNet++~\citep{qi2017pointnet++} & 81.9 & 85.1 & 42.5\\
        \quad Point-BERT~\citep{yu2022pointbert} & 84.1 & 85.6 &  -\\
        \quad Point-MLP~\citep{ma2022rethinking}. & 84.6 & 86.1 & 48.1 \\
        \midrule
        \quad Point-MAE~\citep{yu2022pointbert} & 84.2 & 86.1 & 47.4\\
        \quad M2PT-Video & \reshl{85.6}{1.7\%} & \reshl{87.5}{1.6\%} &\reshl{50.1}{5.7\%}\\
        \quad M2PT-Image & \reshl{85.6}{1.7\%} & \reshl{87.5}{1.6\%} &\reshl{49.2}{3.8\%}\\
        \quad M2PT-Audio & \reshl{85.6}{1.7\%} & \reshl{87.5}{1.6\%} &\reshl{48.1}{1.5\%}\\
        \hline
        \multicolumn{4}{@{\;}l}{\bf From-scratch setting} \\
        \quad N/A & 50.2 & 68.4 & - \\
        \quad M2PT-Video & \reshl{50.8}{1.2\%} & \reshl{68.8}{0.6\%} & - \\
        \bottomrule
        \end{tabular}
        }
\end{table}

 \begin{table}[t]
		\caption{\textbf{Experimental results on AudioSet-2k}. The relative improvements over the baselines are shown in \color{mygreen}{green}.}
  \vspace{-0.1in}
  \label{tab:audio}
		\centering
  \resizebox{0.85\linewidth}{!}{
  \small
  \begin{tabular}{lcc}
                \toprule
				Method & Model &  Top-1 Acc. (\%) \\ \hline
    \multicolumn{3}{@{\;}l}{\bf Pretrained setting} \\
                \quad PSLA~\citep{gong2021psla} & CNN+Trans & 31.9 \\
                \quad AST~\citep{gong2021ast} & ViT-B & 34.7 \\
                \quad SSAST~\citep{gong2022ssast} & ViT-B & 31.0 \\
                \midrule
                \quad AudioMAE~\citep{huang2022masked} & ViT-B & 35.3 \\
                \quad M2PT-Point & ViT-B & \reshl{35.6}{0.8\%} \\
                \quad M2PT-Video & ViT-B & \reshl{35.5}{0.6\%} \\
                \quad M2PT-Image & ViT-B & \reshl{35.6}{0.8\%} \\
                \hline
    \multicolumn{3}{@{\;}l}{\bf From-scratch setting} \\
                \quad N/A & ViT-B &   11.0 \\
                \quad M2PT-Point    &   ViT-B   &   \reshl{11.4}{3.6\%} \\
                \bottomrule
			\end{tabular}
   }
	\end{table}

    \begin{table}[t]
    \centering
    \captionof{table}{\textbf{Experimental results on Kinetics-400}. The relative improvements over the baselines are shown in \color{mygreen}{green}}
    \vspace{-0.1in}
    \label{tab:video}
        \resizebox{0.85\linewidth}{!}{
        \small
            \begin{tabular}{lcc}
                \toprule
                Method & Model & Top-1 Acc. (\%) \\ \hline
                SlowFast-101~\citep{feichtenhofer2019slowfast} & ResNet-101 & 79.8 \\
                MViTv2-B~\citep{li2022mvitv2} & ViT-B & 81.2 \\
                TimeSFormer~\citep{bertasius2021space} & ViT-B & 80.7 \\
                \midrule
                VideoMAE~\citep{tong2022videomae} & ViT-B & 81.5 \\
                M2PT-Point & ViT-B & \reshl{82.3}{1.0\%} \\
                M2PT-Image & ViT-B & \reshl{82.2}{0.9\%} \\
                M2PT-Audio & ViT-B & \reshl{82.3}{1.0\%} \\
                \bottomrule
				\end{tabular}
			}
        \end{table}
    
\noindent\textbf{Video understanding.} For the experiments on Kinetics-400, we adopt only the pretrained setting because it is not a common practice to train a model from scratch on a video dataset, which would deliver inferior performance. We use the Video-MAE-pretrained ViT to initialize the target weights. Naturally, the baseline should be the VideoMAE-pretrained model directly finetuned on Kinetics-400. Table~\ref{tab:video} shows that compared with previous works including SlowFast~\citep{feichtenhofer2019slowfast}, MViTv2~\citep{li2022mvitv2}, TimeSFormer~\citep{bertasius2021space}, and VideoMAE~\citep{tong2022videomae}, M2PT outperforms by at least +0.8 top-1 accuracy (82.3 vs. 81.5), which reveals that the temporal awareness for video understanding can also be enhanced with irrelevant data from other modalities.

\subsection{Ablation Studies} ~\label{sec:exp:ablation_study}
As shown in Table~\ref{tab:ablate}, we evaluate the design choices of M2PT separately through a group of ablation studies under the pretrained setting on ImageNet and the auxiliary modality is the point cloud. We make the following observations.

\noindent\textbf{1) Applying Cross-Modal Re-parameterization to every linear layer delivers the best performance}. In each transformer block, we may choose to apply our method to any of the Query/Key/Value/projection layers in the attention block and the two linear layers in the FFN. Table~\ref{tab:ablate} shows changing any one of the layers brings improvements, and the best result is achieved by changing them all.

    \begin{table*}[t]
   	\centering
   	\caption{\textbf{Ablation studies} on design choices of M2PT including the layers to re-parameterize and configurations of Cross-Modal Scale $\lambda$. We use the point cloud and video as auxiliary modalities for image and 3D evaluation. The first row reports the results of direct tuning.}
   	\label{tab:ablate}
    \small
   		\begin{tabular}{cccc|cc|ccc}
   			\toprule
   			\multicolumn{4}{c|}{Multimodal Pathway Components} & \multicolumn{2}{c|}{Cross-Modal Scale} & \multirow{1}{*}{ImageNet} & \multirow{1}{*}{ShapeNetPart } & \multirow{1}{*}{PartNet}\\
   			\cline{1-4} 
   			\texttt{Attn QKV} & \texttt{Attn Proj} & \texttt{FFN 1st} & \texttt{FFN 2nd}     &  Init.   &  Trainable &(\%) & (\%) &  (\%)  \\ 
   			\hline 
            & & & & - & - & 83.3 & 84.2/86.1 & 47.4\\
   			\ding{52} &    &   &   &   0   &   \ding{52}  &   83.4 & 84.6/86.5 & 48.3   \\
   			& \ding{52}    &   &   &   0   &   \ding{52}  &   83.6  & 84.8/87.1 & 48.2  \\
   			&    & \ding{52}   &   &   0   &   \ding{52}  &   83.6  & 84.9/87.0 & 48.4\\
   			&    &   & \ding{52}   &   0   &   \ding{52}  &   83.7 & 85.2/87.2 & 48.3 \\
   			\ding{52} & \ding{52}   & \ding{52}  &  \ding{52} &   0   &   \ding{52}  &  \textbf{83.9}  & \textbf{85.6}/\textbf{87.5} & \textbf{50.1}\\
   			\hline
   			\ding{52} & \ding{52}   & \ding{52}  &  \ding{52} &   $10^{-2}$  &  \ding{56}   &   83.5  & 84.6/86.3 & 48.2\\
			\ding{52} & \ding{52}   & \ding{52}  &  \ding{52} &   $10^{-2}$   &   \ding{52}  &   83.6  & 84.3/86.2 & 48.0\\
			\ding{52} & \ding{52}   & \ding{52}  &  \ding{52} &   $10^{-4}$   &   \ding{52}  &   83.6  & 84.7/86.2 & 48.1\\
   			\ding{52} & \ding{52}   & \ding{52}  &  \ding{52} &   $10^{-6}$   &   \ding{52}  &   83.7  & 84.7/86.4 & 48.2\\
   			\bottomrule
   	\end{tabular}
    \vspace{-2mm}
   \end{table*}

\noindent\textbf{2) Cross-Modal Scale should be initialized with 0}. By default, we initialize the Cross-Modal Scale $\lambda$ with 0 for every layer. We observe that initializing it to a higher value degrades the performance, suggesting that the initial state of the M2PT should be identical to the target weights (\ie, the weights pretrained with MAE, in this case).

\noindent\textbf{3) Cross-Modal Scale should be learnable}. Fixing the Cross-Modal Scale degrades the performance, suggesting it is important to let the model learn how to combine the target weights and the corresponding auxiliary weights.

    \begin{table}
    \captionof{table}{ImageNet accuracy with changed order of auxiliary weights or fewer pretraining epochs.}
    \label{tab:investigation}
        \small
        \centering
        \begin{tabular}{ccc}
        \toprule
         Order of aux weights & Epochs pretrained & Top-1 acc \\
        \hline 
        Normal      &   20 & 83.55 \\
        Normal      &   220 & 83.69\\
        Normal      &   300 & 83.93\\
        \hline
        Reversed    &   300 & 83.61\\
        \bottomrule
        \end{tabular}
        \vspace{-2mm}
    \end{table}

 \begin{table}[ht]
     \caption{ Training efficiency of Multimodal Pathway.}
    \centering
    \resizebox{0.94\linewidth}{!}{
    \begin{tabular}{lcccc}
    \hline
    Model & Train Time & Train Param. & Inference Time  & Inference Param. \\ \hline
    MAE & 16.95 Hours & 86.3M & 11.64 ms & 86.3M \\
    M2PT & 22.84 Hours & 172.6M & 11.64ms & 86.3M \\ \hline
    \end{tabular}
    }
    \vspace{-4mm}
    \end{table}
\subsection{Empirical Discussions}

\subsubsection{On the Modality-Complementary Knowledge}

The observed improvements on multiple modalities have shown that the auxiliary transformer has learned some knowledge that can be transferred to the target modality. We continue to investigate the properties of such modality-complementary knowledge through two groups of experiments (Table~\ref{tab:investigation}). 

\noindent\textbf{1) Modality-complementary knowledge \& Abstraction Hierarchy}. Vision Transformers excel in general hierarchical representations by stacking blocks~\cite{dosovitskiy2020image}. For example, in the image and point cloud modalities, this hierarchy may include textures (in images) or individual points (in point clouds), object parts, and whole objects. In Table~\ref{tab:investigation}, we construct the multimodal pathway by connecting transformer blocks of different depths. Specifically, the counterpart of the first target block should be the first auxiliary block. Under the reverse-order setting, we observe that doing so decreases the accuracy to 83.61\%, which is 0.32\% lower than the normal M2PT. We observe that modality-complementary knowledge in the auxiliary transformer can transfer to another modality but can be harmed if the low-to-high correspondence is interrupted, suggesting that modality-complementary knowledge reinforces hierarchical representations of the transformer architecture.  

\noindent\textbf{2) More trainable parameters? Just better initialization?} For this part, we use insufficiently pretrained auxiliary weights. Specifically, the default auxiliary weights are pretrained for 300 epochs with mask modeling on point cloud data, but we alternatively use the checkpoints saved at the 20th and 220th epoch, respectively, as the auxiliary weights. Not surprisingly, we observe that the performance degrades to 83.55\% and 83.69\%, respectively, which is still higher than the baseline. This phenomenon suggests that the improvements brought by the auxiliary weights cannot be explained as better initialization, because after pretraining the auxiliary model from 20 to 300 epochs, the accuracy increases from
83.5 to 83.9. If improvements were due to initialization, the results of pretraining 20 epochs should be close to random initialization (83.5 \textit{v.s.} 81.9).

\subsubsection{Discussion on the Data Scale}

\noindent\textbf{1) From small-scale data to large-scale data.} Previous works such as Image2Point~\citep{xu2022image2point} introduces image-pretrained models to data-insufficient 3D perception tasks. Differently, M2PT sets up a brand new methodology and breaks the former consensus - we discover that \emph{even though the data scale of point clouds is limited, such data still brings out impressive improvements to the image, video, and audio perception tasks}. Impressively, the pretraining data of the latter modalities is larger in magnitude than that of the point cloud, but the point cloud data makes a difference. \noindent\textbf{2) From large-scale data to small-scale data.} On the other hand, the effectiveness of M2PT highlights that for 3D vision research and other areas that lack large-scale data for pretraining, M2PT introduces a promising direction to leverage irrelevant large-scale data from other modalities.

\section{Conclusion and Limitation} ~\label{sec:conclusion}
This paper explores the feasibility and advantages of improving a transformer's performance on a specific modality with irrelevant data from other modalities. We propose the Multimodal Pathway and a concrete implementation of no additional inference cost named Cross-Modal Re-parameterization. It represents an early exploration in this direction, which offers a novel perspective. We realize significant and consistent improvements on four representative modalities, demonstrating the potential of our method as a promising approach. In the future, we will explore to construct multimodal pathways among CNNs and cross-architecture. The primary limitation is that the theory behind the improvements remains to be revealed. Apart from empirical explanations, we believe further investigations (\eg, a mathematically provable bound) will be useful.

\noindent{\large\textbf{Acknowledgements.}} This work is partially supported by the National Natural
Science Foundation of China (Grant No. 8326014).

{
    \small
    \bibliographystyle{ieeenat_fullname}
    \bibliography{main}
}

\end{document}